\pdfoutput=1
\documentclass[11pt]{article}

\usepackage{acl}

\usepackage{times}
\usepackage{latexsym}
\usepackage{balance}
\usepackage[T1]{fontenc}
\usepackage{pgfplots} 
\usepackage{amssymb}
\usepackage[utf8]{inputenc}

\usepackage{multirow}
\usepackage{amsmath}
\usepackage{capt-of}
\usepackage{tabularx}
\usepackage{epsfig}
\usepackage{amssymb}
\usepackage{amsfonts}
\usepackage{booktabs}
\usepackage{scalerel}
\usepackage[inline]{enumitem}
\usepackage{listings}
\usepackage{varwidth}
\usepackage[export]{adjustbox}
\usepackage{tikz}
\usetikzlibrary{tikzmark}
\usepackage{cleveref}

\usepackage{stmaryrd}
\usepackage{bbm}

\usepackage{algorithm}
\usepackage[noend]{algpseudocode}

\definecolor{deepblue}{rgb}{0,0,0.5}
\definecolor{officeblue}{RGB}{0,102,204}
\definecolor{deepred}{rgb}{0.6,0,0}
\definecolor{deepgreen}{rgb}{0,0.5,0}
\definecolor{mybrickred}{RGB}{182,50,28}

\definecolor{fillcolor}{RGB}{216,217,252}

\algnewcommand\algorithmicrequireb{{\hspace{0.85cm}}}
\algnewcommand\INPTDESCB{\item[\algorithmicrequireb]}

\algnewcommand\algorithmicfuncdesc{\textbf{Function:}}
\algnewcommand\FUNCDESC{\item[\algorithmicfuncdesc]}
\algnewcommand\algorithmicfuncdescb{{\hspace{1.48cm}}}
\algnewcommand\FUNCDESCB{\item[\algorithmicfuncdescb]}
\algnewcommand{\algorithmicgoto}{\textbf{goto}}
\algnewcommand{\Goto}[1]{\algorithmicgoto~\ref{#1}}

\usepackage{amsmath,amsfonts,bm}

\def\eqref#1{equation~\ref{#1}}

\def\1{\bm{1}}

\def\vtheta{{\bm{\theta}}}
\def\va{{\bm{a}}}

\def\ve{{\bm{e}}}

\def\vg{{\bm{g}}}
\def\vh{{\bm{h}}}

\DeclareMathAlphabet{\mathsfit}{\encodingdefault}{\sfdefault}{m}{sl}
\SetMathAlphabet{\mathsfit}{bold}{\encodingdefault}{\sfdefault}{bx}{n}

\newcommand{\Ls}{\mathcal{L}}

\newcommand\ours{EFT}
\newcommand\oursDRP{\textsc{DPR-r}}

\newcommand{\orsharc}{OR-ShARC}

\title{Bridging The Gap: Entailment Fused-T5 for Open-retrieval Conversational Machine Reading Comprehension }

\author{Xiao Zhang$^{123}$,~~Heyan Huang$^{123}$\thanks{\ \ Corresponding author.},~~Zewen Chi$^{123}$,~~\textbf{Xian-Ling Mao}$^{123}$\\
$^{1}$School of Computer Science and Technology, Beijing Institute of Technology\\
$^2$Beijing Engineering Research Center of High Volume Language Information Processing\\
and Cloud Computing Applications\\
$^3$Southeast Academy of Information Technology, Beijing Institute of Technology\\
\texttt{\{yotta,hhy63,czw,maoxl\}@bit.edu.cn}\\}

\begin{document}
\maketitle
\begin{abstract}
Open-retrieval conversational machine reading comprehension (OCMRC) simulates real-life conversational interaction scenes. Machines are required to make a decision of \texttt{Yes/No/Inquire} or generate a follow-up question when the decision is \texttt{Inquire} based on retrieved rule texts, user scenario, user question, and dialogue history. Recent studies explored the methods to reduce the information gap between decision-making and question generation and thus improve the performance of generation. However, the information gap still exists because these pipeline structures are still limited in decision-making, span extraction, and question rephrasing three stages. Decision-making and generation are reasoning separately, and the entailment reasoning utilized in decision-making is hard to share through all stages. To tackle the above problem, we proposed a novel one-stage end-to-end framework, called Entailment Fused-T5 (\ours{}), to bridge the information gap between decision-making and generation in a global understanding manner. The extensive experimental results demonstrate that our proposed framework achieves new state-of-the-art performance on the OR-ShARC benchmark.
\end{abstract}

\section{Introduction}

\begin{figure}[t]
\centering
\includegraphics[width=0.4\textwidth]{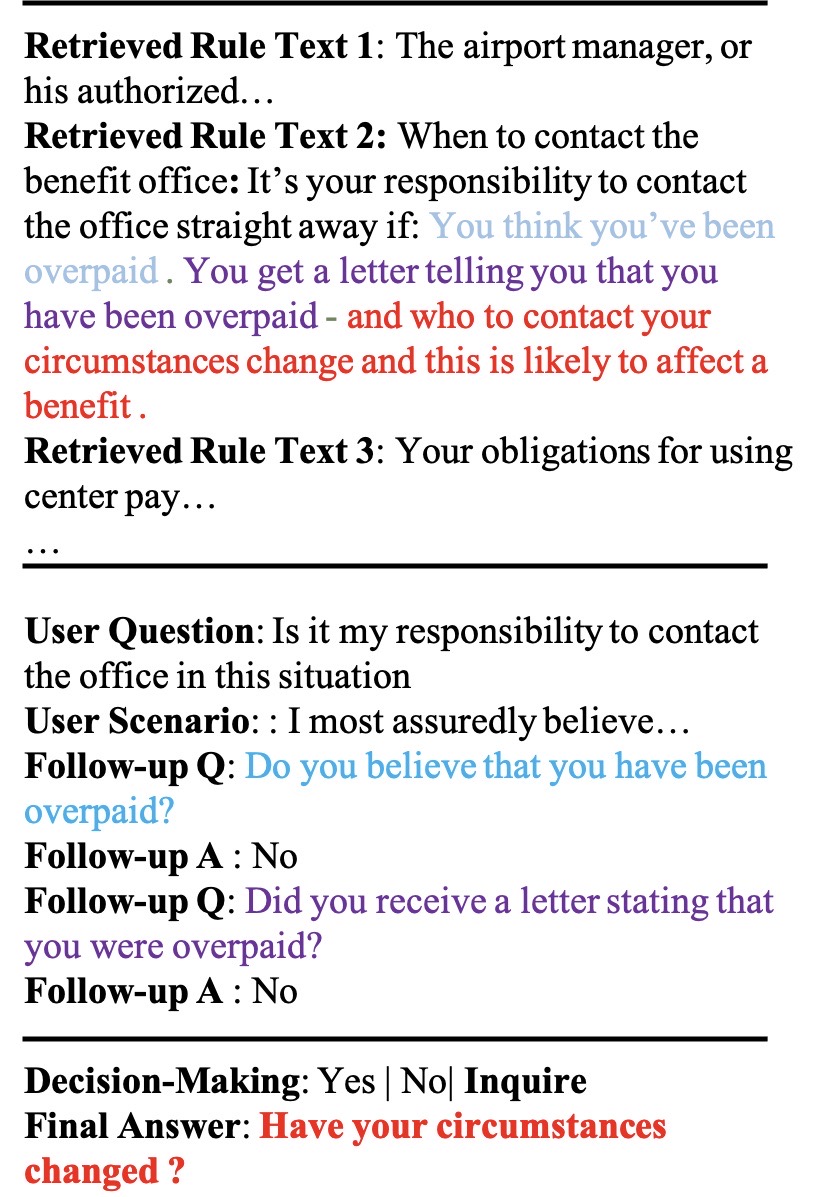}
\caption{An example in the OCMRC dataset. Given the user scenario and user question, machines are required to first retrieve related rule texts in the knowledge database, and then make a decision of \texttt{Yes/No/Inquire} or generate a follow-up question when the decision is \texttt{Inquire} based on retrieved rule texts, user scenario, user question, and dialogue history. }
\label{fig:example}
\end{figure}

Open-retrieval conversational machine reading comprehension (OCMRC) \cite{gao-open-cmrc} investigates real-life scenes, aiming to formulate multi-turn interactions between humans and machines in open-retrieval settings. As shown in Figure \ref{fig:example}, given the user scenario and user question, machines are required to first retrieve related rule texts in the knowledge database, and then make a decision of \texttt{Yes/No/Inquire} or generate a follow-up question when the decision is \texttt{Inquire} based on retrieved rule texts, user scenario, user question, and dialogue history.

\begin{figure*}[t]
\centering
\includegraphics[width=0.99\textwidth]{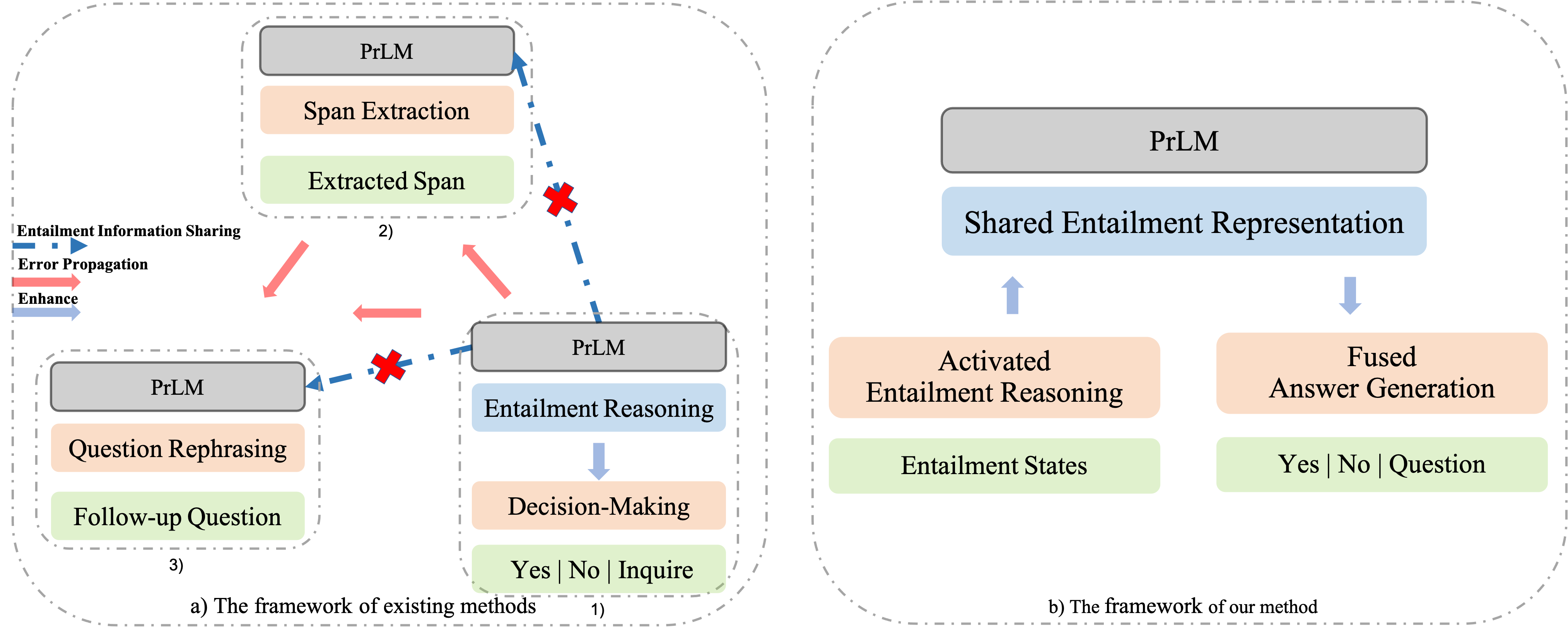}
\caption{The comparison between our framework and previous pipeline framework. a) Previous framework typically has three stages: entailment reasoning based decision-making, span extraction, and question rephrasing. Thus, the entailment reasoning utilized in decision-making is hard to share through all stages. Meanwhile, the performance of previous framework suffers from error propagation problem. The above information misleading leads to the information gap between decision and generation. b) Our framework is a one-stage end-to-end model. To bridge the information gap, the fused answer generation directly generates the decisions or follow-up questions with the shared entailment  representation enhanced by activated entailment reasoning. }
\label{fig:instance}
\end{figure*}

Previous studies \cite{sharc,UrcaNet,lawrence-etal-2019-bison-stage1,zhong-zettlemoyer-2019-e3-span-generation,emt-span-generation,discern-span-generation,graph-span-generation} typically investigate the interactions in a complex context. Entailment reasoning is proposed to explicitly supervise the decision-making classification, and the performance is highly improved. Despite the effectiveness of entailment reasoning, the performance is still limited because of the information gap between decision-making and question generation. Recent studies \cite{smoothing-open-cmrc} explored smooth dialogue methods to reduce the gap between decision-making and question generation.

However, the information gap still exists in OCMRC because these pipeline structures lack global understanding. Specifically, as illustrated in Figure \ref{fig:instance}, nearly all of these methods \cite{smoothing-open-cmrc, gao-open-cmrc} are still limited in decision-making, span extraction, and question rephrasing three stages. Decision-making and question generation are reasoning separately, and the entailment reasoning utilized in decision-making is hard to share through all stages, but suffer from error propagation problem. 
The above information misleading leads to the information gap between decision-making and question generation.

\begin{figure}[t]
\centering
\includegraphics[width=0.4\textwidth]{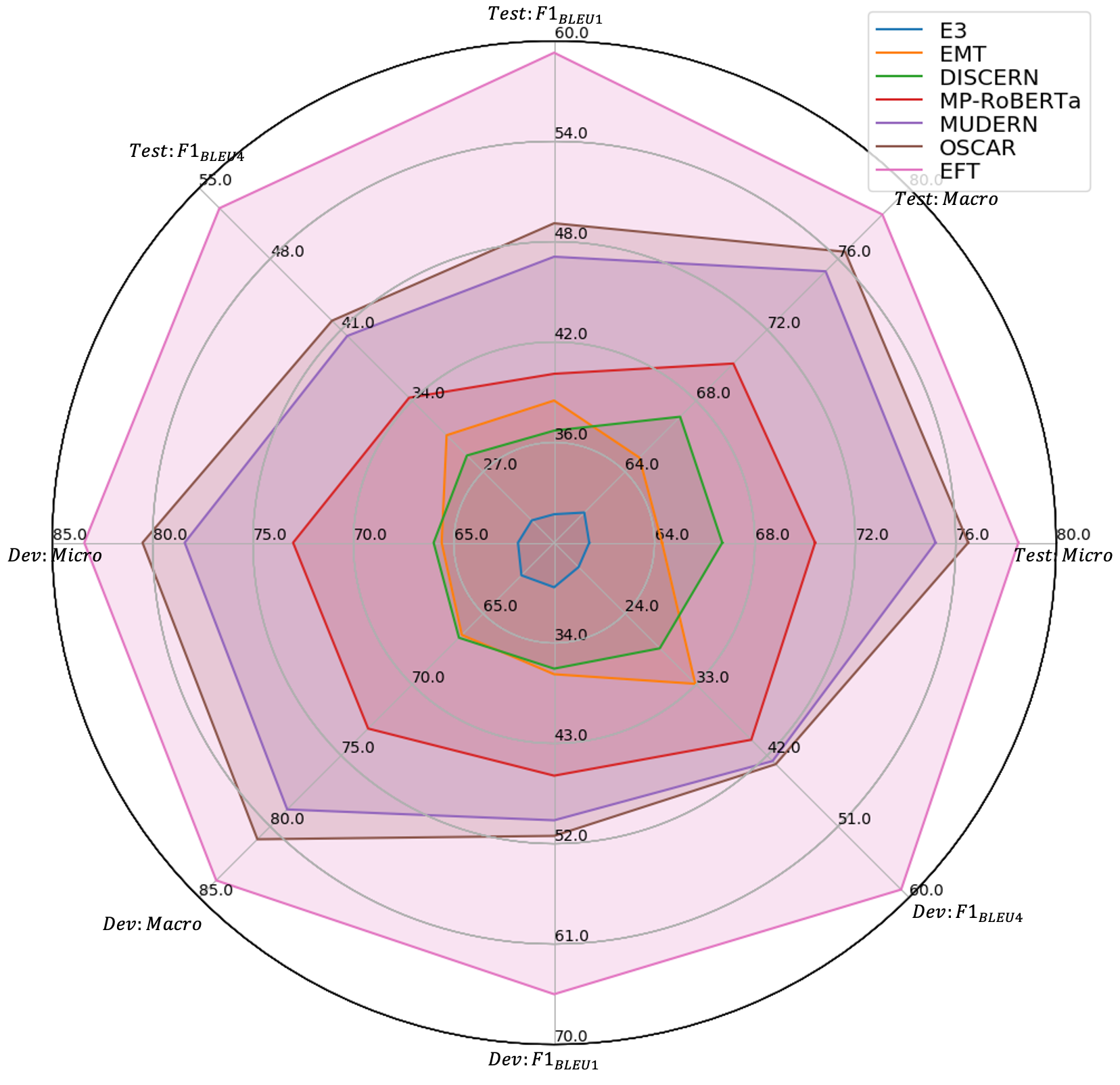}
\caption{ \ours{} has achieves state-of-the-art performance on the OR-ShARC benchmark.}
\label{fig:metric}
\end{figure}

To tackle the above problems, we proposed a novel one-stage end-to-end framework, called entailment fused-T5 (\ours{}) to bridge the information gap between decision-making and question generation in a global understanding manner. Specifically, our model consists of a universal encoder and a duplex decoder. The decoder consists of an activated entailment reasoning decoder and a fused answer generation decoder. The implicit reasoning chains of both decision-making and question generation in the multi-fused answer generation decoder are explicitly supervised by activated entailment reasoning through the shared entailment representation of our encoder. Thus, our model can reason in a global understanding manner. We further investigate a relevance-diversity fusion strategy to improve the implicit reasoning abilities of our model, especially for the implicit ranking among the multiple retrieved rules. The extensive results, as illustrated in Figure \ref{fig:metric}, demonstrate that our proposed framework \ours{} achieves new state-of-the-art performance on the OR-ShARC benchmark.

Our contributions are summarized as follows:
\begin{itemize}
\item We proposed a novel one-stage end-to-end framework, called entailment fused-T5 (\ours{}) to bridge the information gap between decision-making and question generation through a global understanding manner.
\item We further investigate a relevance-diversity fusion strategy (RD strategy) to improve the implicit reasoning abilities of our model.
\item Extensive experiments demonstrate the effectiveness of our proposed framework on the OR-ShARC benchmark.
\end{itemize}

\section{Related Work}

\paragraph{Conversation-based Reading Comprehension}

Conversation-based reading comprehension \cite{sharc,sun-etal-2019-dream,reddy-etal-2019-coqa,choi-etal-2018-quac,cui-etal-2020-mutual,gao-open-cmrc} aims to formulate human-like interactions. Compared to traditional reading comprehension, these tasks extend the reading comprehension scenarios with dialogue interactions.
There are typically three main types of these tasks: span-based QA tasks \cite{choi-etal-2018-quac,reddy-etal-2019-coqa}, multi-choice tasks \cite{sun-etal-2019-dream,cui-etal-2020-mutual}, or hybrid-form tasks  \cite{sharc,gao-open-cmrc}.


\paragraph{Conversational Machine Reading Comprehension}

CMRC \cite{sharc} is the hybrid form of conversation-based reading comprehension, which requires the machines to make a decision or generate a follow-up question based on rule text, user scenario, user question, and dialogue history.

In this paper, we focus on the open-retrieval conversational machine reading (OCMRC) task \cite{gao-open-cmrc}, which further extends the CMRC task into a real-life scenario. Machines are required to first retrieve related rule texts in a knowledge base based on user questions and user scenarios, then machines are required to make a decision of \texttt{Yes/No/Inquire}, or a follow-up question if the decision is \texttt{Inquire} based on the relevant rule texts, user scenario, user question, and dialogue history. 

Due to the hybrid-form task, the previous methods \cite{zhong-zettlemoyer-2019-e3-span-generation,emt-span-generation,discern-span-generation,graph-span-generation, smoothing-open-cmrc} typically adopt pipeline architectures, including decision-making, span extraction, and question phrasing. Various kinds of entailment reasoning strategies are proposed to improve the performance of decision-making.

Despite the effectiveness of entailment reasoning, the performance is still limited because of the information gap between decision-making and question generation. Recent studies \cite{smoothing-open-cmrc,zhang-2022-et5} explored entailment reasoning sharing methods to reduce the gap between decision-making and question generation, but the performance is limited due to its frame flaws. 

In this paper, we proposed a novel one-stage end-to-end model, called entailment fused-T5 (\ours{}) consists of an encoder and a duplex decoder. The details are written in the next sections.

\section{Methods}

\begin{figure*}[t]
\centering
\includegraphics[width=0.99\textwidth]{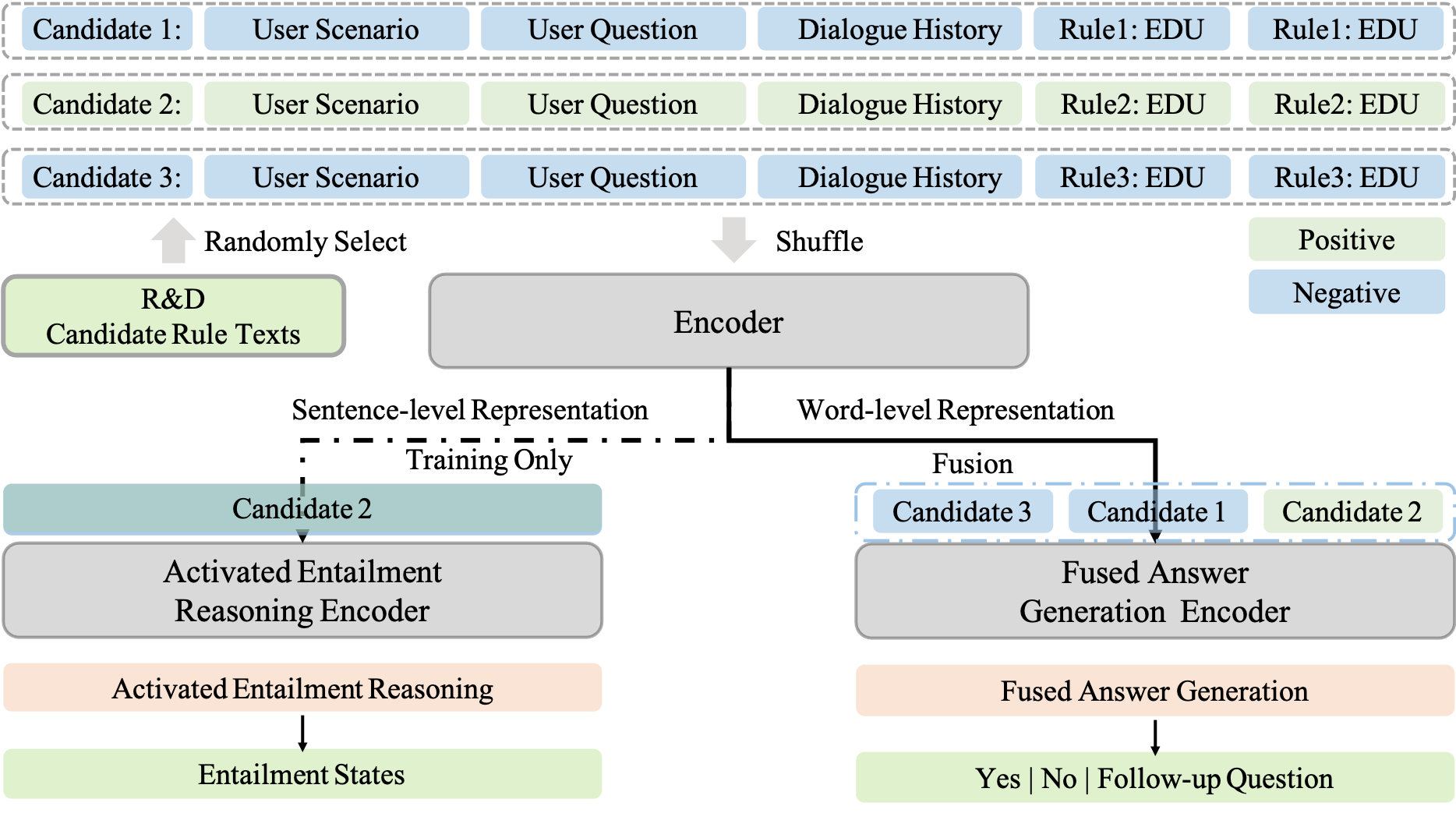}
\caption{The architecture of our proposed \ours{}. Machines first randomly select related rule texts from RD candidate rule texts in the training stage, while in the evaluating stage, machines only use the top-5 retrieved rule texts. Then the input representation is encoded separately, the sentence-level representation is utilized for activated entailment reasoning, and the word-level representation is fused for the final answer generation. }
\label{fig:framework}
\end{figure*}

In open-retrieval CMRC, the machines are first required to retrieve related rule texts in a knowledge base, given user question and user scenario. Then machines are required to make decisions or generate follow-up questions based on retrieved rule texts, user scenario, user question, and dialogue history. Thus, we conduct a retriever to first retrieve related rule texts from the knowledge database, and then generate the final answer through our end-to-end reader \ours{}. The training procedure is shown in Algorithm~\ref{alg:trainP}.


\subsection{Retriever}
We first concatenate the user question and user scenario as the query to retrieve related rule texts in the knowledge base. The knowledge base is divided into the seen subset and the unseen subset. This is to simulate the real usage scenario: users will ask questions about rules they have already seen, or rules that are completely new. We only use seen rules in the training process. In this work, we utilize DPR to retrieve related rule texts. Different previous works typically utilize TF-IDF negatives as DRP hard negatives and limit the data space of retrieved negatives, we random sample rule texts from seen knowledge base as the negatives. Each step will randomly sample $m$ numbers negatives in the training stage. We retrieve the top 20 relevant rule texts for each query, which is further used by our reader.

\subsection{Reader: \ours{}}
In this stage, each item is formed as the tuple $\left \{ R,S,Q,D \right \}$. $R$ donates the  rule text candidates. $R = \left \{ r_{1},r_{2},...,r_{k} \right \}$, where $r$ donates the rule text item of $R$. $S$ and $Q$ represent user scenario and user questions, respectively.   $D$ donates the dialogue history. Given $\left \{ R,S,Q,D \right \}$, \ours{} will directly generate a decision of \texttt{Yes/No/Inquire} or follow-up question when the decision is \texttt{Inquire}.
\ours{} consists of a universal encoder and a duplex decoder. The Duplex decoder consists of an activated entailment reasoning decoder and a fused answer generation decoder. In this way, the whole implicit reasoning chains of the fused answer generation decoder will be fine-grained supervised by activated entailment reasoning with the shared entailment  representation. Thus, the fused answer generation decoder will reason in a global understanding manner. The details are shown in Figure \ref{fig:framework}.

\subsection{Encoding}
Given $\left \{ R,S,Q,D \right \}$, we random sample $k$ items in $R$, and concatenate each of them in $R$ with $S,Q,D$ as $c$, thus the item collection is formed as $C = \left \{ c_{1},c_{2},...,c_{k} \right \}$.  Specifically, each $r$ in $R$ is first parsed to elementary discourse unit (EDU) by a pre-trained model \cite{li2018segbotParser}. The final input format is shown in Figure \ref{fig:framework}. To prevent the leakage of location information in the fused answer generation decoder, and enhance the information extraction ability of the decoder, we utilize a relevance-diversity fusion (RD) fusion strategy to randomly shuffle the order of items which are sampled from RD candidate rule texts, the details are written in Sec \ref{sec:fused}. Given $C$, we utilize T5 encoder as our backbone to get the representation. The presentation of special token are utilized as sentence-level representation $H_{s} = \left \{ h_{s1},h_{s2},...,h_{sk} \right \}$ for activated entailment reasoning decoding. The word-level representation $H_{w} = \left \{ h_{w1},h_{w2},...,h_{wk} \right \}$ are utilized for fused answer generation decoding. 

\subsection{Decoding}

We utilize duplex decoding to explicitly supervise our answer generation stage, which introduced the explicit entailment reasoning information in implicit answer generation reasoning. The answer generation will either generate a decision of \texttt{Yes/No/Inquire} or a follow-up question when the decision is  \texttt{Inquire}. The activated entailment reasoning decoder will reason the entailment states of the EDUs. The duplex decoder is trained in a multi-task form. And the activated entailment reasoning only activates in training stage.

\paragraph{Activated Entailment Reasoning}
Each EDU  will be classified into one of three entailment states, including \texttt{ENTAILMENT}, \texttt{CONTRADICTION}, and \texttt{NEUTRAL}. To get the noisy supervision signals of entailment states, we adopt a heuristic approach\footnote{The noisy supervision signal is a heuristic label obtained by the minimum edit distance.}. This is proposed to simulate fulfillment prediction of conditions in multi-sentence entailment reasoning, which can explicitly supervise the implicit reasoning chains of the answer generation. 

Previous studies typically introduce entailment reasoning in all rule text segmented EDUs. This will greatly increase the proportion of \texttt{NEUTRAL} labels and affect the model effect, because nearly all of the entailment states of EDUs in retrieved irrelevant rules are \texttt{NEUTRAL}, and introducing more noise in the training stage. In our method, entailment reasoning will only activate for the golden rule text. Utilizing this setting, one is to balance the categories of entailment reasoning, and the other is to supervise the implicit reasoning of the fused decoder, which can help the fused decoder infer correct rule text from multiple retrieved rule texts. \label{sec:fused}

Given the sentence-level representation $H_{s}$, we utilize inter-attention reasoning to fully interact with various information in $r$, including EDUs, user question, user scenario, and dialogue history. We utilize an inter-sentence Transformer \cite{devlin-etal-2019-bert} to get the interacted sentence-level representation $G_{s}$. Then, we use a linear transformation to track the entailment reasoning states of each EDU in activate rule text.

\begin{algorithm}[t]
\caption{Training procedure of \ours{}}
\label{alg:trainP}
\begin{algorithmic}[1]
\Require Contextualized context $C$, learning rate $\tau$, 
\Ensure Final answer $A$, activated entailment reasoning state $E$, \ours{} encoder parameters $\vtheta_{e}$, \ours{} fused answer generation decoder parameters $\vtheta_{a}$, \ours{} activated entailment reasoning decoder parameters $\vtheta_{d}$
\State Initialize $\vtheta_{e}$,$\vtheta_{a}$,$\vtheta_{d}$
\While{not converged}
\For {$i = 1,2,\dots,N$}
\State $\vh_{si}, \vh_{wi} = f(c_i, \vtheta_{e}) $ ~~s.t. $\forall c \in \mathcal{C}$
\State $\ve_i = f(h_{si},\vtheta_{d})$ 
\State $\va_i = f(h_{wi},\vtheta_{a})$

\EndFor
\State \textbf{end for}
\State $\vg \gets \nabla_\vtheta \Ls_{}$
\State $\vtheta_{e} \gets \vtheta_{e} - \tau \vg$
\State $\vtheta_{d} \gets \vtheta_{d} - \tau \vg$
\State $\vtheta_{a} \gets \vtheta_{a} - \tau \vg$

\EndWhile
\State \textbf{end while}
\end{algorithmic}
\end{algorithm}

\begin{equation}
e_{i} = W_{c} \tilde{h}_{e_i} + b_{c} \in \mathcal{R}^{3},
\end{equation}
where the $W_c$ is trainable parameters, $e_{i}$ is the predicted score for the three labels of the $i$-th states.

\paragraph{Fused Answer Generation}

Given the word-level representation $H_{w} = \left \{ h_{w1},h_{w2},...,h_{wk} \right \}$ of $R$, we concatenate the $H_{w}$ as the fused representation $f_{w}$. In this manner, our answer generation decoder can reason through all the $k$ items through an implicit ranking mechanism. It is worth mentioning that each item of $H_{w}$ is first fully interacted among rule text, user question, user scenarios, and dialogue history without other multi-rule noise through our encoder.

To improve the information implicit reasoning abilities of our model, we further investigate the relevance-diversity fusion strategy (RD fusion strategy), which consists of relevance-diversity candidate rule texts, order information protection and fused answer generation. 
The rule text candidates are consists of top $k$ relevant rule texts and randomly sampled rule texts, which are called RD candidate rule texts. Thus, the candidates are full-filled with relevant and diverse rule texts. On the premise of ensuring relevance among the rule texts, the diversity of learnable information sampling combinations is further improved. Moreover, the order of items fused in $f_{w}$ may lead to information leakage and affect the reasoning ability of the decoder in the training stage, so as we mentioned in the last section, we will randomly shuffle the order of items when inputting to the encoder to protect the order information. In the evaluation stage, only the top 5 unshuffled relevant rule texts will be utilized for answer generation.

The fused answer generation is utilized to generate either the decision or the follow-up question. We employ T5 decoder as our answer generation decoder. Given encoder fused representation $f_{w}$, and the final answer $a$,  including decision or follow-up question, the answer is composed of the variable-length tokens $x_{i}$, the probabilities over the tokens are shown in the blow:

    \begin{equation}
        p(a) = \prod_{1}^{m}p(x_{i}|x_{<i},f_{e};\theta ),
    \end{equation}
where $\theta$ donates the trainable parameters of our decoder.

\subsection{Training Objective}

\paragraph{Activated Entailment Reasoning}
 The activated entailment reasoning is supervised by cross-entropy loss, by given the entailment stages $c_{i}$ :
\begin{equation}
    \mathcal{L}_{enatil} = - \frac{1}{N} \sum_{i=1}^{N} log \, \mathrm{softmax} (c_i)_{r},
\end{equation}
where $r$ is the ground truth of entailment state.

\paragraph{Fused Answer Generation}
The fused answer generation training objective is computed as illustrated in below:
    \begin{equation}
        \mathcal{L}_{answer} = -\sum_{i=1}^{M} log \, p(x_{i}|x_{<i},f_{w};\theta ),
    \end{equation}
The overall loss function is:
\begin{equation}
    \mathcal{L} = \mathcal{L}_{answer} + \lambda \mathcal{L}_{entail}.
\end{equation}

\section{Experiment and Analysis}
\subsection{Data}
\paragraph{Dataset}
The experiment dataset is OR-ShARC \cite{gao-open-cmrc}, the current OCMRC benchmark. The corpus is clawed from the government website. There is a total of 651 rule texts collected in the knowledge base. For the validation and test set, the golden rule texts are split into seen or seen. This is to simulate the real usage scenario: users will ask questions about rules they have already seen, or rules that are completely new. The train, dev, and test size is 17,936, 1,105, and 2,373, respectively. Each item consists of utterance id, tree id, golden rule document id, user question, user scenario, dialog history, evidence, and the decision.


\subsection{Setup}
\begin{table*}
\centering
\scalebox{0.99}{
\begin{tabular}{lcccccccc}
\toprule
\multirow{3}{*}{Model} &
\multicolumn{4}{c}{Dev Set} & \multicolumn{4}{c}{Test Set}\\
&\multicolumn{2}{c}{Decision Making} & \multicolumn{2}{c}{Question Gen.} & \multicolumn{2}{c}{Decision Making} & \multicolumn{2}{c}{Question Gen.}\\
\cmidrule{2-5}
\cmidrule{6-9}
 & Micro & Macro & $\text{F1}_{\text{BLEU1}}$ & $\text{F1}_{\text{BLEU4}}$ & Micro & Macro & $\text{F1}_{\text{BLEU1}}$ & $\text{F1}_{\text{BLEU4}}$ \\ 
\midrule
 E$^3$ & 61.8\scriptsize{$\pm$0.9} & 62.3\scriptsize{$\pm$1.0} & 29.0\scriptsize{$\pm$1.2} & 18.1\scriptsize{$\pm$1.0}&  61.4\scriptsize{$\pm$2.2} & 61.7\scriptsize{$\pm$1.9} & 31.7\scriptsize{$\pm$0.8} & 22.2\scriptsize{$\pm$1.1}\\
 EMT & 65.6\scriptsize{$\pm$1.6} & 66.5\scriptsize{$\pm$1.5} & 36.8\scriptsize{$\pm$1.1}& 32.9\scriptsize{$\pm$1.1}&  64.3\scriptsize{$\pm$0.5} & 64.8\scriptsize{$\pm$0.4} & 38.5\scriptsize{$\pm$0.5} & 30.6\scriptsize{$\pm$0.4} \\
 DISCERN & 66.0\scriptsize{$\pm$1.6} & 66.7\scriptsize{$\pm$1.8} & 36.3\scriptsize{$\pm$1.9} & 28.4\scriptsize{$\pm$2.1} &  66.7\scriptsize{$\pm$1.1} & 67.1\scriptsize{$\pm$1.2} & 36.7\scriptsize{$\pm$1.4} & 28.6\scriptsize{$\pm$1.2} \\
 MP-RoBERTa & 73.0\scriptsize{$\pm$1.7} & 73.1\scriptsize{$\pm$1.6} & 45.9\scriptsize{$\pm$1.1} & 40.0\scriptsize{$\pm$0.9} &  70.4\scriptsize{$\pm$1.5} & 70.1\scriptsize{$\pm$1.4} & 40.1\scriptsize{$\pm$1.6} & 34.3\scriptsize{$\pm$1.5} \\
 MUDERN &  78.4\scriptsize{$\pm$0.5} & 78.8\scriptsize{$\pm$0.6} &
 49.9\scriptsize{$\pm$0.8} &42.7\scriptsize{$\pm$0.8}\scriptsize{} & 75.2\scriptsize{$\pm$1.0} & 
 75.3\scriptsize{$\pm$0.9} & 47.1\scriptsize{$\pm$1.7}&40.4\scriptsize{$\pm$1.8}\\
OSCAR & 80.5\scriptsize{$\pm$0.5} & 80.9\scriptsize{$\pm$0.6}  & 51.3\scriptsize{$\pm$0.8} & 43.1\scriptsize{$\pm$0.8}& 76.5\scriptsize{$\pm$0.5} & 76.4\scriptsize{$\pm$0.4} & 49.1\scriptsize{$\pm$1.1} & 41.9\scriptsize{$\pm$1.8} \\
\ours{} & \textbf{83.4}\scriptsize{$\pm$0.5} & \textbf{83.8}\scriptsize{$\pm$0.5} &\textbf{65.5}\scriptsize{$\pm$1.9} & \textbf{59.0}\scriptsize{$\pm$2.0}\scriptsize{} & \textbf{78.5}\scriptsize{$\pm$0.7} & \textbf{78.5}\scriptsize{$\pm$0.7} & \textbf{59.3}\scriptsize{$\pm$0.8}& \textbf{53.0}\scriptsize{$\pm$0.8}\\
\bottomrule
\end{tabular}
}
\caption{Results on the dev and test set of OR-ShARC. The average results with standard deviation on 5 random seeds are reported.}\label{table:or-main}
\end{table*}

\begin{table}
\centering
\scalebox{0.8}{
\begin{tabular}{lcccc}
\toprule
 Model & Micro & Macro & $\text{F1}_{\text{BLEU1}}$ & $\text{F1}_{\text{BLEU4}}$ \\ 
\midrule
\ours{} & \textbf{83.4}\scriptsize{$\pm$0.5} & \textbf{83.8}\scriptsize{$\pm$0.5} & \textbf{65.5}\scriptsize{$\pm$1.9} & \textbf{59.0}\scriptsize{$\pm$2.0}\scriptsize{}\\
 \quad -w/o s & 82.9\scriptsize{$\pm$0.6} & 83.4\scriptsize{$\pm$0.5} &63.8\scriptsize{$\pm$1.6} & 57.0\scriptsize{$\pm$1.8}\scriptsize{} \\
 \quad -w/o s+a & 80.7\scriptsize{$\pm$0.8} & 81.1\scriptsize{$\pm$0.9} &62.4\scriptsize{$\pm$2.3} & 56.3\scriptsize{$\pm$2.3}\scriptsize{} \\
 \quad -w/o s+a+i & 80.2\scriptsize{$\pm$0.5} & 80.5\scriptsize{$\pm$0.6} &61.2\scriptsize{$\pm$1.4} & 55.0\scriptsize{$\pm$1.6}\scriptsize{} \\
 \quad -w/o s+a+i+f & 71.0\scriptsize{$\pm$1.2} & 71.6\scriptsize{$\pm$0.9} & 49.2\scriptsize{$\pm$0.8} & 43.8\scriptsize{$\pm$0.8}\scriptsize{} \\
\bottomrule
\end{tabular}
}
\caption{Ablation study of \ours{} on the dev set of OR-ShARC.}\label{table: ablation}
\end{table}

\paragraph{Evaluation}
The evaluation consists of two parts: decision-making and question generation. We utilize Micro-Acc and Macro-Acc for the results of decision-making, and use $\text{F1}_{\text{BLEU}}$ \cite{gao-open-cmrc} for question generation. The $\text{F1}_{\text{BLEU}}$ is conducted to evaluate the question generation performance when the predicted decision is \texttt{Inquire}.
\begin{align}
    precision_{\text{BLEU}} = \frac{\sum_{i=0}^{M} \text{BLEU}(y_i, \hat{y}_i)}{M},
\end{align}
Where $M$ is the total number of \texttt{Inquire} decisions made by our model. $y_i$ is the predicted question, $\hat{y}_i$ is the corresponding ground truth prediction. The recall of BLEU is computed in a similar way.
\begin{align}
    recall_{\text{BLEU}} = \frac{\sum_{i=0}^{N} \text{BLEU}(y_i, \hat{y}_i)}{N},
\end{align}
where $N$ is the total number of \texttt{Inquire} decision from the ground truth annotation, 

The calculation of $\text{F1}_{\text{BLEU}}$ is shown in below:
\begin{align}
    \text{F1}_{\text{BLEU}} = \frac{2 \times precision_{\text{BLEU}} \times recall_{\text{BLEU}}}{precision_{\text{BLEU}} + recall_{\text{BLEU}}}.
\end{align}

\paragraph{Implementation Details}
We utilize the T5-base \cite{t5} as our reader backbone, and additional add an activated entailment reasoning decoder, whose parameters are randomly initialized. We utilize BERT \cite{devlin-etal-2019-bert} as our retriever backbone , whose parameters are initialized from DPR \cite{karpukhin-etal-2020-dense}. For the RD strategy, we use the top-20 retrieved rule texts and 30 randomly sample rule texts as our fused candidates in the training stage, every step we will randomly select 5 samples from the candidates. We only use seen rule texts in the knowledge base for the training stage. And We only use top 5 retrieved rule text for the inference stage. The fused number $k$ is set as 5 for fused answer generation for both training and inference. We use AdamW \cite{loshchilov2018fixing-adamw} to fine-tune our model. The learning rate is hierarchically designed, the learning rate of T5 is 2e-4, and the learning rate of activated entailment decoder is 2e-5. We tried from 0.1 to 1.0 for $\lambda$, and find 0.9 is the best hyper-parameter. The beam-size is set as 5 for the answer generation.
\subsection{Results}
All results on the OR-ShARC benchmark are illustrated in Table \ref{table:or-main}, including dev and test set with metrics for both decision-making and question generation.

Experimental results demonstrate that our proposed methods achieve new SOTA on the OR-ShARC benchmark. \ours{} outperforms OSCAR by 3.6\% in Micro-Acc, 3.6\% in Macro-Acc for decision-making on the dev set, and outperforms OSCAR by 2.6\% in Micro-Acc, 2.7\% in Macro-Acc for decision-making on the test set. In particular, our proposed \ours{} achieves considerable improvement in BLEU scores. \ours{} outperforms OSCAR by 27.7\% in $\text{F1}_{\text{BLEU1}}$, 36.9\% in $\text{F1}_{\text{BLEU4}}$ for the question generation on the dev set,and outperforms OSCAR by 20.8\% in $\text{F1}_{\text{BLEU1}}$, 26.5\% in $\text{F1}_{\text{BLEU4}}$ for the question generation on the test set. We further to investigate the classwise accuracy performance of \ours{}, as shown in Table \ref{tab:class-wise}. Experiments show that the accuracy of each category in OR-ShARC is improved by conducting \ours{} framework, compared with reported baselines.

To further investigate the performance for our proposed \ours{} in seen and unseen settings, the performance of the split subset \footnote{Only BLEU scores are reported in OSCAR.} is illustrated in Table \ref{table: seen-unseen}. Compared with OSCAR, the seen subset performance are greatly improved through our framework \ours{}. \ours{} greatly outperforms OSCAR by 29.1\% in $\text{F1}_{\text{BLEU1}}$, 36.4\% in $\text{F1}_{\text{BLEU4}}$ for the question generation on the seen test set. In addition, compared with the previous pipeline architectures utilized in MURDEN and OSCAR, our model not only improves the performance, but also makes the framework of OCMRC more lightweight. We reduce the number of model parameters from 330M to 220M, which is decreased by 33.3\%. The performance on the seen subset of \ours{} is 35.0\% higher in micro-acc than seen subset, 35.3\% higher in macro-acc than unseen subset. Our retrieval results are illustrated in Table \ref{table:dpr} and Table \ref{table:dpr-split}. The details are illustrated in Appendix~\ref{sec:dpr-r} and Appendix~\ref{sec:subset}, respectively.

\begin{table*}
\small
\setlength{\tabcolsep}{4pt}
\centering\centering
\begin{tabular}{lccccccccc}
\toprule
\multirow{2}{*}{Model} &
\multicolumn{4}{c}{Seen} & \multicolumn{4}{c}{Unseen} & \multirow{2}{*}{Parameters}\\
 & $\text{F1}_{\text{Micro}}$ & $\text{F1}_{\text{Macro}}$  & $\text{F1}_{\text{BLEU1}}$ & $\text{F1}_{\text{BLEU4}}$ &$\text{F1}_{\text{Micro}}$ & $\text{F1}_{\text{Macro}}$  & $\text{F1}_{\text{BLEU1}}$ & $\text{F1}_{\text{BLEU4}}$
 \\ \midrule
  MUDERN &88.1 & 88.1& 62.6  & 57.8 & 66.4&66.0 & 33.1 &24.3 &330M \\
 OSCAR & -- & --  & 64.6  &  59.6 & -- & --  & \textbf{34.9} & \textbf{25.1} &330M \\
\textsc{\ours{}} & \textbf{92.4} & \textbf{92.3} & \textbf{83.4} & \textbf{81.3} & \textbf{68.4} & \textbf{68.2} & \textbf{34.9}& 24.0 & \textbf{220M}\\
\bottomrule
\end{tabular}
\caption{The comparison of question generation on the seen and unseen splits on test-set.}\label{table: seen-unseen}
\end{table*}

\subsection{Ablation Studies}
The ablation studies of \ours{} on the dev set of OR-ShARC benchmark are shown in Table \ref{table: ablation}. There are four ablations of our \ours{} is considered:

\begin{itemize}
\item \textbf{\ours{}-wo/s} trains the model without  relevance-diversity (RD) candidate rule texts. Only top5 randomly shuffled relevant rule texts are considered in the training stage.

\item \textbf{\ours{}-wo/s+a} trains this model additionally remove activated entailment reasoning.

\item \textbf{\ours{}-wo/s+a+i} trains this model further cancels random shuffle in the training stage.

\item \textbf{\ours{}-wo/s+a+i+f} trains this model without multi rule fused answer generation, only top1 retrieved rule text is considered.

\end{itemize}


\paragraph{Analysis of RD Candidate}
We investigate the necessity of the RD candidate rule texts. This strategy is utilized to improve the implicit reasoning abilities of our decoder by improve the learning space of fused candidates. On the premise of ensuring the relevance among the rule texts, the diversity of learnable information sampling combinations is further improved. As shown in Table \ref{table: ablation}, compared with \ours{}, the performance of both decision-making and question generation are dropped when removing RD candidate rule texts, which demonstrate the effective of RD candidate rule texts to improve the information seeking abilities of our fused answer generation decoder. By removing RD candidate rule texts, the mirco-acc is decreased by 0.5, the macro-acc is decrease by 0.4, the $\text{F1}_{\text{BLEU1}}$ is decreased by 1.7, and the $\text{F1}_{\text{BLEU4}}$ is decreased by 2.0. The above results demonstrate the necessity of RD candidate rule texts.

\paragraph{Analysis of Activated Entailment Reasoning}
\ours{}-wo/s+a trains this model additionally remove activated entailment reasoning. As illustrated in Table \ref{table: ablation}, compared with \ours{}-wo/s+a, the performance of both decision-making and question generation are dropped without activated entailment reasoning, the mirco-acc is decreased by 2.2, the macro-acc is decrease by 2.3, the $\text{F1}_{\text{BLEU1}}$ is decreased by 1.4, and the $\text{F1}_{\text{BLEU4}}$ is decreased by 0.7. The above results suggest that the implicit reasoning of conversational machine reading comprehension could be enhanced by introducing explicit fine-grained supervise signal in a global understanding manner.

\paragraph{Analysis of Order Information Protection}
The order of fused representation used in fused answer generation decoder may lead to information leakage and affect the reasoning ability of the decoder. In order to avoid the problem of poor information seeking ability caused by excessive learning of position information of the model, we randomly shuffle the order of fused representation to protect the order information in the training stage. As illustrated in Table \ref{table: ablation}, compared with \ours{}-wo/s+a, \ours{}-wo/s+a+i decrease the performance of both decision-making and question generation without order information protection, the mirco-acc is decreased by 0.5 , the macro-acc is decrease by 0.6, the $\text{F1}_{\text{BLEU1}}$ is decreased by 1.2, and the $\text{F1}_{\text{BLEU4}}$ is decreased by 1.3. The above results indicates the importance of order information protection.

\paragraph{Analysis of Fused Generation}
Fused Generation is utilized to introduce the ability to process multiple rule contextualized information. The multiple rule contextualized information are fused as a single fused information. \ours{}-wo/s+a+i+f trains this model without multi rule fused answer generation, only top1 retrieved rule text is considered. In this manner, the performance is limited with the retrieval performance. Compared with \ours{}-wo/s+a+i+f, the performance of both decision-making and question generation of \ours{}-wo/s+a+i are significantly improved by introducing fused answer generation strategy, the mirco-acc is increased by 9.2 , the macro-acc is increased by 8.9, the $\text{F1}_{\text{BLEU1}}$ is increased by 12.0, and the $\text{F1}_{\text{BLEU4}}$ is increased by 11.2. The above results suggests the necessity of fused answer generation strategy.

\begin{table}
	\centering
\scalebox{0.75} 
{
	\begin{tabular}{lcccccccc}
    \toprule
        \multirow{2}{*}{Model} & \multicolumn{2}{c}{Yes} &  \multicolumn{2}{c}{No} & \multicolumn{2}{c}{Inquire}  \\
        \cmidrule(lr){2-3} \cmidrule(lr){4-5} \cmidrule(lr){6-7} 
        & {dev} & {test} & {dev} & {test} &{dev} & {test}  \\
    \midrule
        E$^3$ & 
        58.5 & 58.5 & 
        61.8 & 60.1 & 
        66.5 & 66.4 \\
        EMT \ & 
        56.9 & 55.4 & 
        68.6 & 65.6 & 
        74.0 & 73.6 \\
        DISCERN \ & 
        61.7 & 65.8 & 
        61.1 & 61.8 & 
        77.3 & 73.6 \\
        MP-RoBERTa\  & 
        68.9 & 72.6 & 
        80.8 & 74.2 & 
        69.5 & 63.4 \\
        MUDERN \ & 
        73.9 & 76.4 & 
        80.8 & 72.2 & 
        81.7 & 77.4 \\
        \ours\ & 
        \textbf{80.1} & \textbf{81.2} & 
        \textbf{83.2} & \textbf{75.6} & 
        \textbf{88.2} & \textbf{78.7} \\
    \bottomrule
    \end{tabular}
    }  \caption{
	   Class-wise decision making accuracy among ``Yes'', ``No'' and ``Inquire'' on the dev and test set of \orsharc. 
	}\label{tab:class-wise}
\end{table}

\section{Conclusion}
In this paper, we propose a novel end-to-end framework, called \ours{}, to bridge the information gap between decision-making and question generation through the shared entailment representation in a global understanding manner. In addition, we utilize the RD fusion strategy for the fused answer generation decoder to handle the multi-rule problem through an implicit ranking method. Extensive experimental results on the OR-ShARC benchmark demonstrate the effectiveness of our proposed framework \ours{}. In our analysis, the implicit reasoning ability of both decision-making and question generation is significantly improved by sharing external explicit entailment knowledge through our novel framework \ours{}. 



\bibliographystyle{acl_natbib}
\bibliography{cc}

\appendix

\begin{table*}
\centering
\scalebox{0.9}
{\begin{tabular}{lcccccccc}
\toprule
\multirow{2}{*}{Model} &
\multicolumn{4}{c}{Dev Set} & \multicolumn{4}{c}{Test Set}\\
 & Top1 & Top5 & Top10 & Top20 & Top1 & Top5 & Top10 & Top20 \\ 
 \midrule
 TF-IDF & 53.8 & 83.4 & 94.0  & 96.6 &  66.9& 90.3 & 94.0 & 96.6 \\
 DPR & 48.1 & 74.6 & 84.9 & 90.5 & 52.4 & 80.3 & 88.9 & 92.6 \\
 TF-IDF + DPR & \textbf{66.3}  & 90.0 & 92.4 & 94.5 & \textbf{79.8} & \textbf{95.4} & \textbf{97.1} & 97.5\\
 \midrule
\oursDRP{}(ours) & 54.5 & \textbf{93.4} & \textbf{99.2} & \textbf{99.5} & 77.5 & 93.5 & 96.3 & \textbf{98.8} \\
  
\bottomrule
\end{tabular}
}
\caption{Comparison of the open-retrieval methods. }\label{table:dpr}
\end{table*}

\begin{table}
\centering
\scalebox{0.95}
{\begin{tabular}{lcccc}
\toprule
\oursDRP{} & Top1 & Top5 & Top10 & Top20 \\ 
 \midrule
Dev & 54.5 & 93.4 & 99.2 & 99.5  \\
\quad Seen Only  & 96.8  & 100.0 & 100.0 & 100.0 \\
\quad Unseen Only & 19.5 & 87.9 & 98.5 & 99.0 \\
Test & 77.5 & 93.5 & 96.3 & 98,8 \\
\quad Seen Only &97.6 & 100.0 & 100.0 & 100.0 \\
\quad Unseen Only & 62.8 & 88.8 & 93.7 & 97.9 \\
\bottomrule
\end{tabular}
}
\caption{Retrieval Results of \oursDRP{}. }\label{table:dpr-split}
\end{table}
\newpage
\section{The Performance of Retriever (\oursDRP{})}\label{sec:dpr-r}
Table~\ref{table:dpr} presents the detailed performance of \oursDRP{}, including the performance of Top-$k$ on dev set and test set. Different from previous methods \cite{smoothing-open-cmrc} that utilize DRP but only use TF-IDF retrieved negatives, we use random negatives sampled from seen rule texts in knowledge base. Experimental results illustrate that \oursDRP{} outperforms DPR by 16.5\% in top5 accuracy on test set, and reaches competitive results with TF-IDF+DPR. 

\section{The Performance of Retriever (\oursDRP{}) on Subset}\label{sec:subset}

We further analysis the performance of \oursDRP{} on the seen and unseen subset. As shown in Table \ref{table:dpr-split}, experimental results demonstrate the effectiveness of \oursDRP{} on seen sets, the top1 accuracy reached 97.6 on the seen subset of test set. But the performance still have a large latent space of improvement on unseen sets. 


\end{document}